\documentclass[runningheads]{llncs}
\usepackage[T1]{fontenc}
\usepackage{graphicx}
\usepackage[misc,geometry]{ifsym}

\usepackage{orcidlink}

\usepackage{adjustbox}

\usepackage{amsfonts}
\usepackage{amsmath}
\usepackage{color}
\usepackage{todonotes}
\usepackage{multirow}

\usepackage[ruled,vlined,linesnumbered,noend]{algorithm2e}

\begin{document}
\title{FairFinGAN: Fairness-aware Synthetic Financial Data Generation}
\titlerunning{FairFinGAN: Fairness-aware Synthetic Financial Data Generation}
%
\author{Tai Le Quy\inst{1}\Letter\orcidlink{0000-0001-8512-5854} \and
Dung Nguyen Tuan \inst{2} \orcidlink{0009-0003-2508-0756}\and
Trung Nguyen Thanh \inst{2} \orcidlink{0009-0005-9847-7038}\and
Duy Tran Cong\inst{2} \and
Huyen Giang Thi Thu \inst{3}\orcidlink{0009-0007-6283-3111}\and
Frank Hopfgartner\inst{1,4}\orcidlink{0000-0003-0380-6088}
}

\authorrunning{T. Le Quy et al.}
%
\institute{University of Koblenz, Germany \\
\email{tailequy@uni-koblenz.de}\\
\and
Hanoi University of Science and Technology, Vietnam\\
\email{\{nguyentuandung23012003, trungnguyenthanh.eworking, congduytran12\}@gmail.com}
\and
Banking Academy of Vietnam, Vietnam\\
\email{huyengtt@hvnh.edu.vn}
\and
University of Sheffield, United Kingdom \\
\email{hopfgartner@uni-koblenz.de}
}

\maketitle              
\begin{abstract}

Financial datasets often suffer from bias that can lead to unfair decision-making in automated systems. In this work, we propose FairFinGAN, a WGAN-based framework designed to generate synthetic financial data while mitigating bias with respect to the protected attribute. Our approach incorporates fairness constraints directly into the training process through a classifier, ensuring that the synthetic data is both fair and preserves utility for downstream predictive tasks. We evaluate our proposed model on five real-world financial datasets and compare it with existing GAN-based data generation methods. Experimental results show that our approach achieves superior fairness metrics without significant loss in data utility, demonstrating its potential as a tool for bias-aware data generation in financial applications.

\keywords{bias \and fairness \and financial data \and GAN \and  synthetic data}
\end{abstract}
\section{Introduction}
\label{sec:intro}
Data play a crucial role in automated decision-making systems within financial organizations. However, accessing and sharing financial data remain difficult and challenging due to privacy concerns. Hence, synthetic data generation offers an efficient solution for the research community \cite{assefa2020generating}, because such generative models provide scalable, privacy-protecting substitutes for datasets that cannot be shared because of sensitivity or ownership constraints \cite{sanchez2025condfairgen}. In addition, machine learning (ML) models used in critical domains such as finance often exhibit bias toward individuals or demographic groups defined by protected attributes, such as gender, race, or age \cite{kozodoi2022fairness,le2022survey}. These biases may come from multiple sources, such as historical discrimination, data collection processes, and algorithmic shortcomings \cite{mehrabi2021survey}. More importantly, concerns have emerged that synthetic data may reproduce or even amplify the underlying biases present in the original data \cite{sanchez2025condfairgen}.

To this end, we aim to address fairness at the data level in the finance domain, i.e., generating fair synthetic data, focusing on tabular data which is the most widely used type in real-world settings. In general, fair data are defined as data that achieve statistical parity with respect to a known protected attribute and the target variable \cite{krchova2023strong,rajabi2022tabfairgan}. In addition, since classifiers trained on fairness-aware generated data are expected to produce fair outcomes \cite{duong2024trusting}, we further evaluate the fairness of classifiers trained on the generated data.

In this paper, we propose a framework for generating fair synthetic financial data using GAN, namely FairFinGAN. Our contributions are as follows: (i) we introduce FairFinGAN, a WGAN-based framework for fairness-aware synthetic financial data generation; (ii) we propose a training strategy that incorporates fairness constraints, specifically statistical parity and equalized odds, into the GAN objective based on a multi-layer perceptron classifier on generated samples, to mitigate bias at the dataset level; (iii) we conduct extensive experiments on five real-world financial datasets, showing that FairFinGAN improves fairness while preserving utility compared to baseline methods.

The rest of our paper is structured as follows. Sect. \ref{sec:relatedwork} gives an overview of the related work. The technical description of the proposed FairFinGAN model is presented in Sect. \ref{sec:FairFinGAN}. Sect. \ref{sec:evaluation} describes the details of our experiments on various datasets. Finally, the conclusion and outlook are summarized in Sect. \ref{sec:conclusion}.

\section{Related Work}
\label{sec:relatedwork}

Earlier methods for generating tabular data relied on statistical models to preserve the joint distribution of mixed-type data \cite{assefa2020generating,xu2019modeling}. Then, the creation of deep-generative models, such as generative adversarial networks (GANs) \cite{goodfellow2020generative}, represents a significant acceleration of the field. For example, MedGAN \cite{choi2017generating} used auto-encoders and GANs for the synthesis of high-dimensional binary medical records, while Table-GAN \cite{park2018data} used convolutional architectures to generate tabular data. CTGAN \cite{xu2019modeling} incorporated mode-specific normalization as well as a conditional generator to handle mixed-type data, imbalanced categorical columns, and multi-modal continuous distributions. 

In responding to the perpetuation of bias in synthetic data, efforts shifted progressively towards table generation of tabular data with fairness. Such generators work towards generations of realistic data and fairness constraint information, e.g., demographic parity, equalized odds. FairGAN \cite{xu2018fairgan} was one of the pioneering GAN-based frameworks to generate fair information. Using an auxiliary discriminator to ensure the independence from protected information, FairGAN was able to suppress disparate impact in the resulting synthetic data. Later, FairGAN+ \cite{xu2019fairgan+} achieves both fair data generation and classification by jointly training the generative model and the classifier through adversarial games involving multiple discriminators. Subsequently, TabFairGAN \cite{rajabi2022tabfairgan} was introduced as an extension of the Wasserstein GAN (WGAN) \cite{arjovsky2017wasserstein} for tabular data, incorporating a fairness penalty in generator loss. TabFairGAN demonstrated improved stability and performance compared to previous GAN-based tabular data generators, and outperformed pre-processing techniques such as reweighting and disparate impact removal on multiple benchmark datasets. More recent methods include information-theoretic and representation-learning approaches, such as information-minimizing GANs \cite{chen2024information}, which minimize mutual information between protected attributes, generated data, and classifier outputs, while simultaneously maximizing both utility and fairness. 

Beyond GAN-based approaches, DECAF \cite{van2021decaf} incorporates structural causal information and applies inference-time debiasing by constraining unbiased edges during data generation. This approach highlights the need to study both the data generation process and the ways in which end-users consume generated data in fairness-aware learning. In addition, PreFair \cite{pujol2023prefair} and FLAI library \cite{gonzalez2024mitigating} employ causal Bayesian networks to generate fair synthetic data through structural and conditional probability modifications of the causal model. They can simultaneously mitigate multiple biases by defining clear, interpretable fairness mechanisms and supporting counterfactual reasoning. 


\section{Proposed Method}
\label{sec:FairFinGAN}

In this section, we present \textit{FairFinGAN}, a WGAN-based fairness-aware synthetic financial data generator, inspired by the TabFairGAN \cite{rajabi2022tabfairgan} approach. The main idea is to employ a second training phase to penalize the discrimination score in the generated data. We use a Multi-Layer Perceptron (MLP) classifier trained on the original dataset to calculate fairness scores on generated samples as a loss component. The network is then iteratively updated based on this loss to improve its fairness performance. We define two versions, namely \textit{FairFinGAN-SP} that uses statistical parity (SP) to score the fairness of the MLP classifier, and \textit{FairFinGAN-EOd}, which scores with the equalized odds (EOd).

We denote $\mathcal{D} = (\mathcal{X},\mathcal{Y},\mathcal{S})$ as a dataset with a set of attributes $\mathcal{X}$, a class labels set $\mathcal{Y} = \{0,1\}$ and protected attributes $\mathcal{S}=\{0,1\}$, i.e., each data point $d^{(i)} = (x, y, s)$. The dataset contains $N_{num}$ numerical and $N_{cat}$ categorical attributes. 

\begin{figure}[h]
\centering
\includegraphics[width=0.8\linewidth]{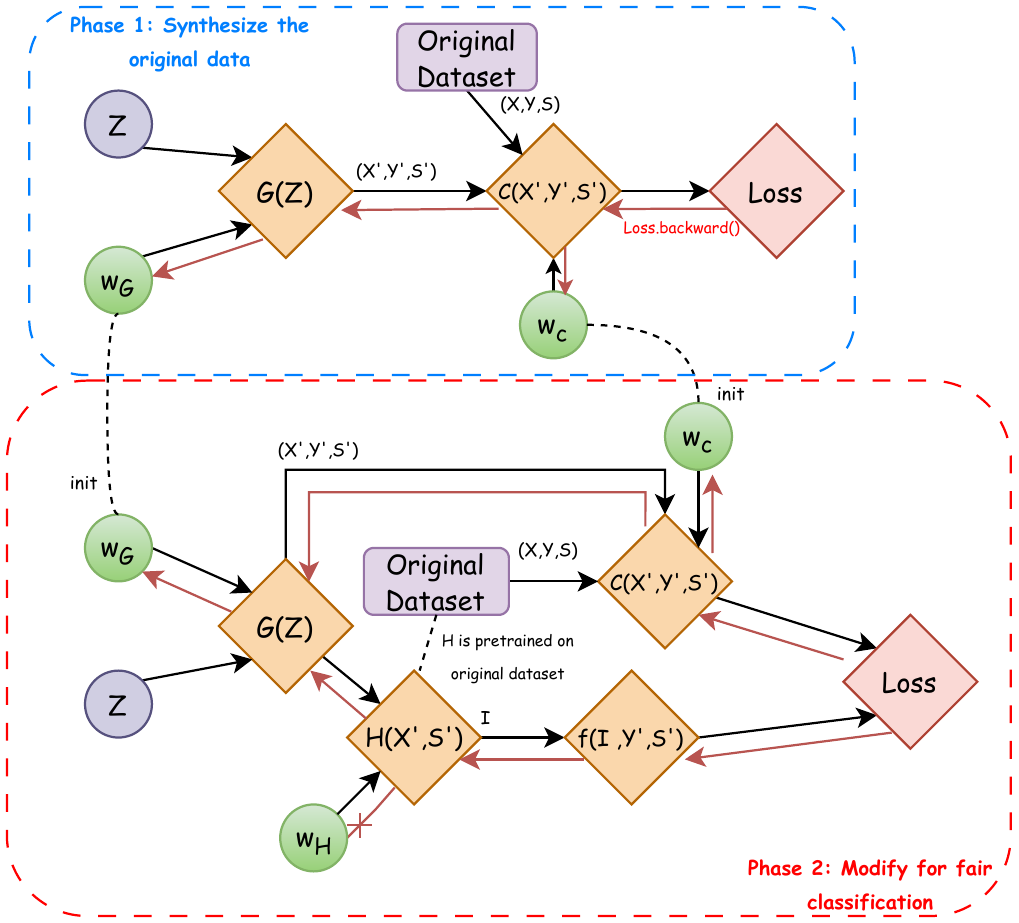}
\caption{Overview of the FairFinGAN model}
\label{fig:pipeline}
\end{figure}

\textbf{The training process.}
The FairFinGAN model training process consists of two phases, as visualized in Fig. \ref{fig:pipeline}. In phase 1, both the generator (G) and the critic (C) participate in the adversarial process, aiming to generate samples as realistically as possible. The generator and critic parameters, denoted by $w_G$ and $w_C$, are updated by the backpropagation method, and the backward propagation is demonstrated by red flows in Fig.\ref{fig:pipeline}.
 Then, in phase 2, an MLP classifier, denoted by $\mathcal{H}$ with parameters $w_\mathcal{H}$, trained on real data $\mathcal{D}$, is used to evaluate the fairness of the classification on the new data generated by the generator after the first phase. The fairness metrics SP and EOd (denoted as $f(I, Y^\prime, S^\prime)$) are then considered as part of the loss function, with a parameter $\lambda_{fair} > 0$. 
 
In particular, each noise vector $z$ is sampled from the normal distribution $P_z$. Let $G(z) = (x^\prime, y^\prime, s^\prime)$ denote one synthetic sample generated by the generator $G$, where $x^\prime$ is the set of attributes, $s^\prime$ is the protected attribute, and $y^\prime$ is the class label. A batch of generated samples is denoted as $(X^\prime, Y^\prime, S^\prime)$, then will be criticized by the critic part of the network. In addition, we use $I = \mathcal{H}(X^\prime, S^\prime)$ to denote the predicted soft label (differentiable) obtained via Gumbel-Softmax on the synthetic data. The formal description of the training process is illustrated in Algorithm \ref{alg:FairFinGAN_training}. Two versions employing different fairness measures (SP and EOd) are depicted in lines \ref{line:SP} and \ref{line:EOd}, respectively.

\begin{algorithm}[!h] 
\SetAlgoLined
\KwIn{Training data $\mathcal{D}$, $total\_epochs$, $fair\_epochs$, $\lambda_{fair}$, $\lambda_{pen}$, $n_{critic}$
}
\KwOut{Generator $G$, critic $C$}

Train a classifier $\mathcal{H}$ on $\mathcal{D}$

\For{$epoch \gets 1$ \KwTo total\_epochs}{
    \If{$epoch \le$ total\_epochs $-$ fair\_epochs}{mode $\gets$ \texttt{accuracy}}
    \Else{mode $\gets$ \texttt{fairness}}

    \ForEach{batch in $\mathcal{D}$}{       
        \For{$k \gets 1$ \KwTo $n_{critic}$}{
        Sample batch $m$ real data $D=(X, Y, S) =\{d^{(1)}, \ldots, d^{(m)}\}$
        
        Sample $m$ noises $Z = \{z^{(1)}, \ldots, z^{(m)}\} \sim P(z)$

        Sample $\varepsilon \sim \mathcal{U}(0,1)$
        
        Generate $D^\prime=(X^\prime, Y^\prime, S^\prime) \gets \{G(z^{(1)}), \ldots, G(z^{(m)})\}$ 

        Compute $\hat{D}=(\hat{X},\hat{Y},\hat{S}) \gets \varepsilon(D)+(1-\varepsilon)(D^\prime)$
        
        Update the critic $C$ by descending the gradient: $\frac{1}{m}\nabla_{w_{C}}\sum_{i=1}^{m}C(D^\prime)-C(D)+ \lambda_{pen}\left(\left\| \nabla_{\hat{D}} C(\hat{D}) \right\|_2 - 1\right)^{2}$
        }

        \If{mode = \texttt{accuracy}}{\tcc{Phase 1. Synthesize the original data}
            Sample $m$ noises $Z = \{z^{(1)},  \ldots, z^{(m)}\} \sim P(z)$

            Update the generator $G$: $\frac{1}{m}\nabla_{w_G}\sum_{i=1}^{m}-(C(G(z^{(i)})))$
            
        }
        \Else{ \tcc{Phase 2.  Modify for fair classification}
            Generate $D^\prime=(X^\prime, Y^\prime, S^\prime) \gets \{G(z^{(1)}), \ldots, G(z^{(m)})\}$

            $I=\{I^{(1)}, \ldots, I^{(m)}\} \gets \{\mathcal{H}(G(z^{(1)})), \ldots, \mathcal{H}(G(z^{(m)}))\}$
            
            Update the generator $G$ (\texttt{FairFinGAN-SP}):
            $\frac{1}{m}\nabla_{w_{G}}\sum_{i=1}^{m} - C(D^\prime) + \lambda_{fair} \left|\frac{\sum_{i: s^\prime_i = 0} I^{(i)}}{\sum_{i:s^\prime_i = 0} 1 } - \frac{\sum_{i: s^\prime_i = 1} I^{(i)}}{\sum_{i:s^\prime_i = 1} 1 }\right|$
            \label{line:SP}
                        
            Or, update the generator $G$ (\texttt{FairFinGAN-EOd}):
            $\frac{1}{m}\nabla_{w_{G}}\sum_{i=1}^{m} - C(D^\prime) + \lambda_{fair}\sum_{y\in\{0,1\}}\left| \frac{\sum_{i: s^\prime_i = 0,\, y^\prime_i = y} I^{(i)}}{\sum_{i: s^\prime_i = 0,\, y^\prime_i = y} 1} - \frac{\sum_{i: s^\prime_i = 1,\, y^\prime_i = y} I^{(i)}}{\sum_{i: s^\prime_i = 1,\, y^\prime_i = y} 1} \right|$ 
            \label{line:EOd}
            
        }
    
    }
}
\Return{$G, C$}\;
\caption{Training process of the FairFinGAN model}
\label{alg:FairFinGAN_training}
\end{algorithm}

\textbf{Neural network architecture.}
The architecture of the generator $G$ is defined as follows:

$ \left\{ \begin{array}{llll}
h_0 = z \\ 
h_1 = \mathrm{ReLu}(FC_{N_{dim}\to N_{dim}}(h_0)) \\
h_2= \mathrm{ReLu}(FC_{N_{dim}\to N_{num}}(h_1)) \bigoplus \mathrm{gumbel}_{0.2}(FC_{N_{dim}\to N_{1}}(h_1))\bigoplus\\ \mathrm{gumbel}_{0.2}(FC_{N_{dim}\to N_{2}}(h_1)) \bigoplus \ldots \bigoplus \mathrm{gumbel}_{0.2}(FC_{N_{dim}\to N_{cat}}(h_1))
\end{array}\right.$

In which, $z$ denotes a latent vector; $FC_{a\to b}$ denotes a fully connected layer with input size $a$ and output size $b$; $\mathrm{ReLu}(v)$  is the \texttt{ReLu} activation applied to the vector input $v$; $\mathrm{gumbel}_{\tau}(v)$ denotes applying \texttt{Gumbel} softmax with parameter $\tau$ on a vector $v$ (in our case $\tau=0.2$); $\bigoplus$ denotes the concatenation of vectors; $N_{dim}$ is the dimension of a transformed data instance, and $N_i$  with $i=1\ldots N_{cat}$ is the dimension of the one-hot encoding vector of the i-th discrete attribute. In addition, the critic $C$ is constructed as follows, where the \texttt{negative\_slope} parameter of the \texttt{LeakyReLU} function is set to its default value of 0.01.

$ \left\{ \begin{array}{lll}
h_0 = v \\ 
h_1 = \mathrm{LeakyReLu}(FC_{N_{dim}\to N_{dim}}(h_0)) \\
h_2= \mathrm{LeakyReLu}(FC_{N_{dim}\to N_{dim}}(h_1)) 
\end{array}\right.$

The classifier $\mathcal{H}$ (MLP) is implemented with two hidden layers and its architecture is defined as follows, where $v$ is an input vector.

$ \left\{ \begin{array}{lll}
h_0 = v \\
h_1 = \mathrm{ReLU}\!\left( FC_{N_{dim}\,\to\,128}(h_0) \right) \\
h_2 = \mathrm{ReLU}\!\left( FC_{128\,\to\,64}(h_1) \right) \\
h_3 = FC_{64,\to\,2}(h_2)
\end{array} \right.$


\section{Evaluation}
\label{sec:evaluation}
In this section, we evaluate our proposed method on five real-world datasets and two state-of-the-art GAN-based data generators. The source code is available at \url{https://github.com/tailequy/FairFinGAN}

\subsection{Experimental Setups}
\label{subsec:setup}

We evaluate our generative methods on five commonly used datasets in the finance and fairness-aware ML domains. All datasets are imbalanced, which is indicated by the imbalance ratio (IR) in Table \ref{tbl:datasets}. Gaussian normalization is applied to the continuous attributes, whereas categorical attributes are encoded using one-hot encoding. The protected attribute ``Age'' in the Credit card and Credit scoring datasets is binarized into $\{25\text{--}65, <25\ \text{or}\>65\}$ following \cite{le2022survey}.

\begin{table}[h!]  
\caption{An overview of financial datasets}
\label{tbl:datasets}
\centering
\begin{adjustbox}{width=\textwidth}
\begin{tabular}{lrrcccc}
\hline
\multicolumn{1}{c}{\textbf{ Datasets }} &  
\multicolumn{1}{c}{\textbf{ \#Instances}} &
\multicolumn{1}{c}{\begin{tabular}[c]{@{}c@{}}\textbf{ \#Instances}\\\textbf{  (cleaned)} \end{tabular}} &
\multicolumn{1}{c}{\textbf{ \#Attributes }} &
\multicolumn{1}{c}{\begin{tabular}[c]{@{}c@{}}\textbf{ Protected }\\\textbf{ attribute(s) } \end{tabular}} & 
\multicolumn{1}{c}{\begin{tabular}[c]{@{}c@{}}\textbf{ Class label}\\\textbf{ (positive) } \end{tabular}} &
\textbf{ IR (+:-) }
 \\ \hline
Adult          & 48,842  & 45,222  & 15  & Gender, Race & $>50K$  & 1:3.03  \\
Credit card    & 30,000  & 30,000  & 24  & Sex, Age     & Default (1)    & 1:3.52   \\
Credit scoring & 8,755   & 8,755   & 17  & Sex, Age     & Good credit (1)      & 11.58:1    \\
Dutch census   & 60,420  & 60,420  & 12  & Sex          & Occupation (1)    & 1:1.10     \\
German credit  & 1,000   & 1,000   & 21  & Sex          & Good credit (1)    & 2.33:1     \\
\hline
\end{tabular}
\end{adjustbox}
\end{table}

We generate a synthetic dataset containing the same number of records as the real-world dataset. To quantify the quality of the synthetic data, we employ four traditional classifiers, including Logistic Regression (LR), Decision Tree (DT), k-nearest neighbors (kNN), and MLP. We use these models because fairness-aware predictive models may make evaluation difficult, as they can self-adjust to produce fairer prediction results. Predictive models are evaluated with the 5-fold cross-validation strategy and average results are reported. The default hyperparameters of these predictive models are used as implemented in the scikit-learn library. Then, we compare our proposed method with the following state-of-the-art GAN-based models: CTGAN \cite{xu2019modeling} and TabFairGAN \cite{rajabi2022tabfairgan}. We use the default hyperparameters of these models as reported in the original papers. The hyperparameters in our proposed models are set as follows: learning rate = 0.0002 (for the generator and critic), learning rate = 0.0001 (for the fairness-aware generator in the second phase), total epochs = 200, fair epochs = 50, batch size = 256, $\lambda_{fair} = 0.5$, $\lambda_{pen} = 10$ and $n_{critic}=4$. 
The generative models were executed in the Kaggle environment (\url{https://www.kaggle.com/}) equipped with a P100 GPU, 16 GB of RAM and 200 GB of storage.

\subsection{Experimental Results}
\label{subsec:results}
At the dataset level, we computed the SP values to assess the fairness of the original and synthetic datasets. The best values are in \textbf{bold}, and the second-best values are in \textit{italic}. Our models achieve the best or second-best values in most datasets with various protected attributes (Table \ref{tbl:sp_dataset}).

\begin{table}[h!]
\caption{The statistical parity on original and synthetic datasets}
\label{tbl:sp_dataset}
\centering
\begin{adjustbox}{width=\textwidth}
\begin{tabular}{ l | c | c| c | c | c }
\hline
\textbf{Dataset-Prot. attr.} &\textbf{Original} &\textbf{CTGAN} &\textbf{TabFairGan} &\textbf{FairFinGAN-SP} &\textbf{FairFinGAN-EOd} \\
\hline

Adult-Gender &0.1989 &0.1474 &\textbf{\textcolor{black}{0.0273}} &\textit{0.1371} &0.1989  \\
Adult-Race &0.1040  & \textit{0.0668} &-0.7533 &-0.6534 &\textbf{\textcolor{black}{0.0043}}  \\
Credit card-Sex &\textit{0.0339}  &-0.0347 & 0.0344  &\textbf{\textcolor{black}{0.0164}} &0.0407  \\
Credit card-Age &-0.0557 &\textit{0.0494} &-0.0966  &\textbf{\textcolor{black}{-0.0081}} &-0.1679  \\
German credit-Sex &0.0748 &0.0235 &-0.0577  &\textbf{\textcolor{black}{0.0118}} &\textit{0.0172}  \\
Credit scoring-Sex &\textit{0.0304}  &0.0854 &0.0597  &-0.0506 &\textbf{\textcolor{black}{0.0017}}  \\
Credit scoring-Age & 0.0752 &\textit{0.0641} &0.1019 &\textbf{\textcolor{black}{0.0338}} &0.0850 \\
Dutch census-Sex & 0.2985  &\textbf{\textcolor{black}{0.2407}} &\textit{0.2708} &0.2919 &0.2873  \\
\hline
\end{tabular}
\end{adjustbox}
\end{table}

Then, we report the performance of four classifiers w.r.t. Accuracy (Acc), Balance accuracy (BA), and seven well-known fairness measures \cite{le2022evaluation}, including SP, equal opportunity (EO), EOd, predictive parity (PP), predictive equality (PE), treatment equality (TE) and Absolute Between-ROC Area (ABROCA). TE can be unbounded because the number of false positives can be zero \cite{le2022evaluation}.
In the Adult dataset (protected attribute ``Gender''), FairFinGAN achieves a reasonable trade-off between fairness and accuracy. Both variants obtain Acc and BA values slightly lower than CTGAN but higher than TabFairGAN (Table \ref{tbl:adult_gender}). The fairness (SP, ABROCA) is generally better than that of CTGAN, though not as extreme as TabFairGAN which shows a good fairness but poor predictive performance. Among them, FairFinGAN-SP achieves slightly better fairness in the LR classifier, while FairFinGAN-EOd shows a comparative result in the kNN model. For the protected attribute ``Race''  (Table \ref{tbl:adult_race}), although both FairFinGAN variants show mixed results, some fairness measures (SP, EO, EOd, PP) are improved with FairFinGAN-EOd (kNN and DT), but our models are better than TabFairGAN in terms of the trade-off between fairness and accuracy.

\begin{table}[h!]
\caption{Adult: performance of predictive models. Protected attribute: Gender}
\label{tbl:adult_gender}
\centering
\begin{adjustbox}{width=\textwidth}
\begin{tabular}{l|l|c|c|c|c|c|c|c|c|c}
\hline
\textbf{Classifier} & \textbf{Method} & \textbf{Acc} & \textbf{BA} & \textbf{SP} & \textbf{EO} & \textbf{EOd} & \textbf{PP} & \textbf{PE} & \textbf{TE} & \textbf{ABROCA} \\
\hline
\multirow{5}{*}{MLP}
& Original &\textit{0.7742}&\textbf{\textcolor{black}{0.5571}}&0.0242&0.0221&0.0262&\textbf{\textcolor{black}{0.0808}}&0.0041&-37.0920&0.0147\\
& CTGAN &\textbf{\textcolor{black}{0.8123}}&\textit{0.5104}&0.0047&\textit{0.0105}&\textit{0.0107}&0.2799&\textit{0.0003}&\textbf{\textcolor{black}{-25.5417}}&0.0131\\
& TabFairGan & 0.7193&0.5&\textbf{\textcolor{black}{0.0}} &\textbf{\textcolor{black}{0.0}} &\textbf{\textcolor{black}{0.0}} &nan&\textbf{\textcolor{black}{0.0}} &nan&\textbf{\textcolor{black}{0.0001}} \\
& FairFinGAN-SP &0.7492&0.5071&-0.0042&0.0180&0.0211&0.2662&0.0032&-107.4718&\textit{0.0084}\\
& FairFinGAN-EOd &0.7392&0.5088&-\textit{0.0010}&0.0218&0.0232&\textit{0.1743}&0.0014&-\textit{28.1228}&0.0104\\
\hline
\multirow{5}{*}{kNN}
& Original &\textit{0.7718}&\textbf{\textcolor{black}{0.6228}}&0.0689&\textit{0.0147}&\textit{0.0369}&0.2504&0.0222&-\textit{2.0808}&\textit{0.0159}\\
& CTGAN &\textbf{\textcolor{black}{0.8127}}&\textit{0.5832}&0.0343&0.0415&0.0496&0.2751&\textit{0.0081}&-3.2464&0.0234\\
& TabFairGan &0.6855&0.5438&0.0157&\textbf{\textcolor{black}{0.0125}} &\textbf{\textcolor{black}{0.0265}} &\textbf{\textcolor{black}{0.0195}} &0.0140&\textbf{\textcolor{black}{-0.0382}} &\textbf{\textcolor{black}{0.0140}}\\
& FairFinGAN-SP &0.7213&0.5454&-\textit{0.0091}&0.0781&0.0869&\textit{0.1171}&0.0088&-2.2987&0.0583\\
& FairFinGAN-EOd &0.7146&0.5635&\textbf{\textcolor{black}{0.0077}} &0.0749&0.0818&0.2185&\textbf{\textcolor{black}{0.0069}}&-2.4996&0.0398\\
\hline
\multirow{5}{*}{DT}
& Original &\textit{0.8094}&\textbf{\textcolor{black}{0.7491}}&0.1989&0.0974&0.2034&0.114&0.1060&-\textit{0.0552}&0.0433\\
& CTGAN &\textbf{\textcolor{black}{0.8180}}&0.7123&0.1438&0.1390&0.2144&0.1627&\textit{0.0754}&-0.0596&0.0503\\
& TabFairGan &0.7561&0.6983&\textbf{\textcolor{black}{0.0265}} &\textbf{\textcolor{black}{0.0172}} &\textbf{\textcolor{black}{0.0389}} &\textbf{\textcolor{black}{0.0158}} &\textbf{\textcolor{black}{0.0217}} &\textbf{\textcolor{black}{-0.004}} &\textbf{\textcolor{black}{0.0149}} \\
& FairFinGAN-SP &0.7710&0.6978&\textit{0.1321}&\textit{0.0514}&\textit{0.1381}&\textit{0.0697}&0.0867&-0.0570&\textit{0.0313}\\
& FairFinGAN-EOd &0.7774&\textit{0.7139}&0.1898&0.0874&0.2021&0.1155&0.1147&-0.1062&0.0372\\
\hline
\multirow{5}{*}{LR}
& Original &\textit{0.7878}&\textbf{\textcolor{black}{0.6194}}&0.0609&\textbf{\textcolor{black}{0.0195}} &\textit{0.0341}&0.2295&0.0146&-\textit{3.7596}&\textit{0.0270}\\
& CTGAN &\textbf{\textcolor{black}{0.8191}}&0.5603&0.0242&\textit{0.0222}&\textbf{\textcolor{black}{0.0278}} &0.2214&\textit{0.0057}&-7.1932&\textbf{\textcolor{black}{0.0189}} \\
& TabFairGan &0.7314&0.5592&0.0348&0.0446&0.0713&\textit{0.0798}&0.0267&5.0814&0.0461\\
& FairFinGAN-SP &0.7483&0.5701&\textbf{\textcolor{black}{0.0010}} &0.1164&0.1242&\textbf{\textcolor{black}{0.0577}} &0.0078&\textbf{\textcolor{black}{-2.8856}}&0.0884\\
& FairFinGAN-EOd &0.7447&\textit{0.5718}&\textit{0.0130}&0.0946&0.0997&0.1802&\textbf{\textcolor{black}{0.0051}}&-4.8309&0.0584\\
\hline
\end{tabular}
\end{adjustbox}
\end{table}

\begin{table}[h!]
\caption{Adult: performance of predictive models. Protected attribute: Race}
\label{tbl:adult_race}
\begin{adjustbox}{width=\textwidth}
\begin{tabular}{l|l|c|c|c|c|c|c|c|c|c}
\hline
\textbf{Classifier} & \textbf{Method} & \textbf{Acc} & \textbf{BA} & \textbf{SP} & \textbf{EO} & \textbf{EOd} & \textbf{PP} & \textbf{PE} & \textbf{TE} & \textbf{ABROCA} \\
\hline
\multirow{5}{*}{MLP}
& Original &\textit{0.7864}&\textbf{\textcolor{black}{0.5986}} &0.0267&0.0149&0.0212&\textbf{\textcolor{black}{0.0774}} &0.0063&-\textit{2.5285}&0.0085\\
& CTGAN &\textbf{\textcolor{black}{0.8126}}&\textit{0.5114}&0.0022&0.0067&0.0075&0.1444&0.0009&-19.7778&0.0104\\
& TabFairGan &0.7298&0.5001&\textbf{\textcolor{black}{0.0001}}&\textbf{\textcolor{black}{0.0002}}&\textbf{\textcolor{black}{0.0003}}&nan&\textbf{\textcolor{black}{0.0}}&nan&\textbf{\textcolor{black}{0.0018}}\\
& FairFinGAN-SP &0.7523&0.5055&-0.0723&0.0713&0.1101&0.2013&0.0388&-32.0556&0.018\\
& FairFinGAN-EOd &0.7463&0.5063&-\textit{0.0013}&\textit{0.0062}&\textit{0.0063}&\textit{0.0806}&\textit{0.0001}&\textbf{\textcolor{black}{-0.4435}}&\textit{0.0058}\\
\hline
\multirow{5}{*}{kNN}
& Original &\textit{0.7718}&\textbf{\textcolor{black}{0.6228}}&0.0414&\textit{0.0303}&\textit{0.0462}&\textit{0.1255}&0.0159&-\textit{0.898}&\textbf{\textcolor{black}{0.0185}} \\
& CTGAN &\textbf{\textcolor{black}{0.8127}}&\textit{0.5832}&0.0159&0.0567&0.0612&0.1729&\textbf{\textcolor{black}{0.0045}}&-1.6905&0.0372\\
& TabFairGan &0.7043&0.5556&-\textit{0.137}&0.0614&0.1792&0.6318&0.1206&nan&0.1714\\
& FairFinGAN-SP &0.7283&0.5503&-0.4425&0.4022&0.7826&0.5512&0.3804&3.5177&0.2073\\
& FairFinGAN-EOd &0.7191&0.5623&\textbf{\textcolor{black}{0.0016}} &\textbf{\textcolor{black}{0.0257}}&\textbf{\textcolor{black}{0.0350}} &\textbf{\textcolor{black}{0.0416}} &\textit{0.0092}&\textbf{\textcolor{black}{0.0571}} &\textit{0.0340}\\
\hline
\multirow{5}{*}{DT}
& Original &\textit{0.8099}&\textbf{\textcolor{black}{0.7498}}&0.0877&\textit{0.0324}&\textit{0.0692}&\textit{0.0905}&0.0368&-0.1993&\textit{0.0203}\\
& CTGAN &\textbf{\textcolor{black}{0.8170}}&0.7127&\textit{0.0504}&0.1214&0.1280&0.1665&\textbf{\textcolor{black}{0.0065}} &-0.159&0.0592\\
& TabFairGan &0.7730&\textit{0.7130}&-0.7518&0.4816&1.3313&0.485&0.8433&-\textit{0.1034}&0.1809\\
& FairFinGAN-SP &0.7668&0.6901&-0.5776&0.3270&0.7277&0.4058&0.4007&1.9027&0.1203\\
& FairFinGAN-EOd &0.7710&0.7001&\textbf{\textcolor{black}{0.0087}}&\textbf{\textcolor{black}{0.0235}} &\textbf{\textcolor{black}{0.0344}} &\textbf{\textcolor{black}{0.0126}} &\textit{0.0109}&\textbf{\textcolor{black}{0.0382}} &\textbf{\textcolor{black}{0.0153}} \\
\hline
\multirow{5}{*}{LR}
& Original &\textit{0.7878}&\textbf{\textcolor{black}{0.6194}}&\textit{0.0155}&\textit{0.0452}&\textit{0.0561}&\textit{0.1467}&\textit{0.0109}&-2.9489&\textbf{\textcolor{black}{0.0367}}\\
& CTGAN &\textbf{\textcolor{black}{0.8191}}&\textit{0.5603}&\textbf{\textcolor{black}{0.0018}}&\textbf{\textcolor{black}{0.0344}}&\textbf{\textcolor{black}{0.0443}}&0.2392&\textbf{\textcolor{black}{0.0098}}&-7.3517&\textit{0.0451}\\
& TabFairGan &0.7265&0.5498&-0.1174&0.0485&0.1044&0.5678&0.0656&nan&0.1982\\
& FairFinGAN-SP &0.7584&0.5544&-0.6167&0.5800&1.0027&0.4223&0.4227&-\textit{0.9967}&0.2218\\
& FairFinGAN-EOd &0.7455&0.5431&-0.0243&0.0699&0.0826&\textbf{\textcolor{black}{0.0767}} &0.0127&\textbf{\textcolor{black}{0.0387}} &0.0485\\
\hline
\end{tabular}
\end{adjustbox}
\end{table}

As demonstrated in Table \ref{tbl:creditcard_sex} (Credit card dataset, protected attribute “Sex”), FairFinGAN-SP achieves the highest accuracy in most predictive models, while FairFinGAN-EOd often attains the best or second-best values for fairness metrics such as EO and EOd, particularly for classifiers DT and LR.
TabFairGan has good results in the case of MLP classifier but suffers from missing value in TE. For the protected attribute ``Age'' (Table \ref{tbl:creditcard_age}), FairFinGAN-EOd achieves the best or second-best fairness measures (EO, EOd, PP, and ABROCA) with the MLP and LR classifier, while the baseline TabFairGan shows strong performance in accuracy for MLP and DT, as well as moderate quality w.r.t. fairness.

\begin{table}[h!]
\caption{Credit card: performance of predictive models. Protected attribute: Sex}
\label{tbl:creditcard_sex}
\begin{adjustbox}{width=\textwidth}
\begin{tabular}{l|l|c|c|c|c|c|c|c|c|c}
\hline
\textbf{Classifier} & \textbf{Method} & \textbf{Acc} & \textbf{BA} & \textbf{SP} & \textbf{EO} & \textbf{EOd} & \textbf{PP} & \textbf{PE} & \textbf{TE} & \textbf{ABROCA} \\
\hline
\multirow{5}{*}{MLP}
& Original &\textit{0.7786} &0.5 &0.0002&\textit{0.0008}& 0.0011&\textit{0.2500}&0.0003& \textit{72.0834} & 0.0014\\
& CTGAN &0.7727&\textit{0.5005}&-0.0003&0.003&0.0041&0.4833&0.0011&\textbf{\textcolor{black}{-34.8333}}&0.0027\\
& TabFairGan &0.7708&0.5003&\textbf{\textcolor{black}{-0.0001}}&\textbf{\textcolor{black}{0.0004}} &\textbf{\textcolor{black}{0.0004}} &\textbf{\textcolor{black}{0.0}} &\textbf{\textcolor{black}{0.0}} &nan&\textbf{\textcolor{black}{0.0009}} \\
& FairFinGAN-SP &\textbf{\textcolor{black}{0.7829}}&\textbf{\textcolor{black}{0.5009}}&0.0002&0.0014&0.0019&0.4667&0.0005&nan&\textit{0.0011}\\
& FairFinGAN-EOd &0.7765&0.5002&-\textit{0.0001} &0.0009&\textit{0.0010}&0.8334&\textit{0.0001}&167.0000&0.0044\\
\hline
\multirow{5}{*}{kNN}
& Original &0.7560&0.5506&\textit{0.0140}&\textit{0.0163}&\textit{0.0251}&0.0465&\textbf{\textcolor{black}{0.0088}}&-0.2803&\textbf{\textcolor{black}{0.0101}} \\
& CTGAN &0.7453&0.5589&\textbf{\textcolor{black}{-0.0109}}&\textbf{\textcolor{black}{0.0091}} &\textbf{\textcolor{black}{0.0205}} &\textit{0.0244}&\textit{0.0114}&\textit{0.2514}&0.0124\\
& TabFairGan &\textit{0.8054}&\textbf{\textcolor{black}{0.6683}} &0.0869&0.1226&0.1835&0.0613&0.0608&1.7007&0.0177\\
& FairFinGAN-SP &\textbf{\textcolor{black}{0.8095}}&\textit{0.6547}&0.0199&0.0264&0.0423&\textbf{\textcolor{black}{0.0229}} &0.0159&0.3308&0.0234\\
& FairFinGAN-EOd &0.7766&0.5926&0.0249&0.0327&0.0455&0.0472&0.0128&\textbf{\textcolor{black}{-0.0479}} &\textit{0.0102}\\
\hline
\multirow{5}{*}{DT}
& Original &0.7246&0.613&0.0377&0.0478&0.0774&0.0343&0.0297&\textbf{\textcolor{black}{-0.0132}} &\textit{0.0199}\\
& CTGAN &0.7559&0.6599&\textbf{\textcolor{black}{-0.0227}}&\textit{0.0443}&\textit{0.0534}&0.0686&\textbf{\textcolor{black}{0.0091}} &0.0908&0.0213\\
& TabFairGan &\textbf{\textcolor{black}{0.8006}} &\textbf{\textcolor{black}{0.7204}} &0.0637&0.0607&0.1055&\textit{0.0228}&0.0448&0.2708&0.0247\\
& FairFinGAN-SP &\textit{0.7951}&\textit{0.704}&0.0371&0.0678&0.0877&\textbf{\textcolor{black}{0.0186}} &0.0199&0.1847&0.0263\\
& FairFinGAN-EOd &0.7673&0.6717&\textit{0.0330}&\textbf{\textcolor{black}{0.0334}}&\textbf{\textcolor{black}{0.0500}}&0.0508&\textit{0.0166}&-\textit{0.0637}&\textbf{\textcolor{black}{0.0128}} \\
\hline
\multirow{5}{*}{LR}
& Original &0.7788&0.5001&\textbf{\textcolor{black}{-0.0001}} &\textbf{\textcolor{black}{0.0005}} &\textbf{\textcolor{black}{0.0007}} &1.0&\textbf{\textcolor{black}{0.0002}} &nan&\textit{0.0185}\\
& CTGAN &0.7773&0.5368&-0.0142&0.0288&0.0354&0.0609&0.0067&\textbf{\textcolor{black}{-1.3176}}&0.0222\\
& TabFairGan &\textit{0.8085} &\textit{0.6148}&0.0572&0.1121&0.1421&0.0858&0.0300&9.6583 &\textbf{\textcolor{black}{0.0164}} \\
& FairFinGAN-SP &\textbf{\textcolor{black}{0.8236}}&\textbf{\textcolor{black}{0.6394}} &0.0332&0.0941&0.1051&\textbf{\textcolor{black}{0.0242}} &0.0109&\textit{1.7570}&0.0607\\
& FairFinGAN-EOd &0.8064&0.5963&\textit{0.0094}&\textit{0.0158}&\textit{0.0187} &\textit{0.0516}&\textit{0.0029}&-1.9989&0.0198\\
\hline
\end{tabular}
\end{adjustbox}
\end{table}

\begin{table}[h!]
\caption{Credit card: performance of predictive models. Protected attribute: Age}
\label{tbl:creditcard_age}
\begin{adjustbox}{width=\textwidth}
\begin{tabular}{l|l|c|c|c|c|c|c|c|c|c}
\hline
\textbf{Classifier} & \textbf{Method} & \textbf{Acc} & \textbf{BA} & \textbf{SP} & \textbf{EO} & \textbf{EOd} & \textbf{PP} & \textbf{PE} & \textbf{TE} & \textbf{ABROCA} \\
\hline
\multirow{5}{*}{MLP}
& Original &0.7787&0.5001&0.0004&0.0005&0.0008&nan&\textit{0.0003}&nan&\textit{0.0019}\\
& CTGAN &0.7727&\textbf{\textcolor{black}{0.5009}}&0.0008&0.0034&0.0045&0.4318&0.0011&-122.2619&0.0140\\
& TabFairGan &\textbf{\textcolor{black}{0.7837}}&0.5&\textit{0.0003}&\textit{0.0003}&\textit{0.0006}&nan&\textbf{0.0002}&nan&\textbf{\textcolor{black}{0.0007}} \\
& FairFinGAN-SP &0.7797&0.5&\textbf{\textcolor{black}{0.0002}}&\textit{0.0003}&\textbf{\textcolor{black}{0.0005}} &nan&\textbf{\textcolor{black}{0.0002}} &nan&0.0042\\
& FairFinGAN-EOd &0.7770&0.5&\textbf{\textcolor{black}{-0.0002}} &\textbf{\textcolor{black}{0.0002}} &\textit{0.0006}&\textbf{\textcolor{black}{0.2}}&0.0004&\textbf{\textcolor{black}{-68.0}}&0.0059\\
\hline
\multirow{5}{*}{kNN}
& Original &0.7560&0.5506&-0.0582&0.0683&0.1153&\textit{0.0524}&0.0469&\textit{-0.6795}&\textit{0.0349}\\
& CTGAN &0.7453&0.5589&\textbf{\textcolor{black}{0.0086}} &0.0462&\textbf{\textcolor{black}{0.0555}}&0.1244&\textbf{\textcolor{black}{0.0093}} &\textbf{\textcolor{black}{-0.6202}} &0.0411\\
& TabFairGan &\textit{0.7771}&\textit{0.5918}&\textit{-0.0306}&\textit{0.0413}&\textit{0.0681}&\textbf{\textcolor{black}{0.0516}}&\textit{0.0268}&0.8689&0.0617\\
& FairFinGAN-SP &0.7565&0.5619&-0.0401&\textbf{\textcolor{black}{0.0380}}&0.0901&0.1081&0.0520&-0.9165&0.0598\\
& FairFinGAN-EOd &\textbf{\textcolor{black}{0.7786}}&\textbf{\textcolor{black}{0.5934}} &-0.0809&0.0881&0.1172&0.1896&0.0291&1.3108&\textbf{\textcolor{black}{0.0183}} \\
\hline
\multirow{5}{*}{DT}
& Original &0.7238&0.612&-0.0730&\textbf{\textcolor{black}{0.0398}} &0.1084&\textit{0.0333}&0.0685&\textbf{\textcolor{black}{-0.0720}}&\textbf{\textcolor{black}{0.0252}}\\
& CTGAN &0.7554&0.6606&\textbf{\textcolor{black}{0.0285}}&0.0690&\textbf{\textcolor{black}{0.0950}}&0.0690&\textbf{\textcolor{black}{0.0261}}&-0.1515&\textit{0.0320}\\
& TabFairGan &\textbf{\textcolor{black}{0.7801}} &\textbf{\textcolor{black}{0.6811}} &-0.0783&\textit{0.0401}&\textit{0.1049}&\textbf{\textcolor{black}{0.0330}}&0.0647&\textit{0.1508}&0.0412\\
& FairFinGAN-SP &\textit{0.7647}&\textit{0.6664}&\textit{-0.0667}&0.0997&0.1534&0.0372&\textit{0.0536}&-0.3464&0.0348\\
& FairFinGAN-EOd &0.7588&0.6583&-0.1396&0.1134&0.1862&0.1646&0.0728&0.2177&0.0345\\
\hline
\multirow{5}{*}{LR}
& Original &0.7788&0.5001&\textbf{\textcolor{black}{0.0001}} &\textbf{\textcolor{black}{0.0003}} &\textbf{\textcolor{black}{0.0004}} &nan&\textbf{\textcolor{black}{0.0001}} &nan&\textit{0.0269}\\
& CTGAN &0.7773&0.5368&0.0206&0.0470&0.0596&\textit{0.1265}&\textit{0.0126}&7.5732&0.0427\\
& TabFairGan &\textbf{\textcolor{black}{0.8078}}&\textbf{\textcolor{black}{0.6075}} &\textit{-0.0002}&\textit{0.0466}&0.0631&0.1456&0.0165&11.7463&0.1055\\
& FairFinGAN-SP &\textit{0.7987}&\textit{0.5791}&-0.0706&0.1286&0.1810&0.1390&0.0525&\textit{-7.2820}&0.0426\\
& FairFinGAN-EOd &0.7915&0.5604&-0.0394&0.0337&\textit{0.0467}&\textbf{\textcolor{black}{0.1135}}&0.0129&\textbf{\textcolor{black}{4.1103}}&\textbf{\textcolor{black}{0.0244}} \\
\hline
\end{tabular}
\end{adjustbox}
\end{table}

In the Credit scoring dataset (protected attribute ``Sex''), FairFinGAN-EOd achieves the best or second-best fairness results (SP, EO, EOd, PP, and TE) while maintaining competitive accuracy (Table \ref{tbl:creditscoring_sex}) across all predictive models. Moreover, FairFinGAN-SP shows balanced performance with high accuracy and improved fairness (SP, protected attribute ``Age'') in several classifiers (MLP and LR). FairFinGAN-EOd achieves competitive fairness in some models but with a minor trade-off in accuracy. TabFairGan performs well in accuracy, but tends to exhibit weaker fairness across most fairness measures (Table \ref{tbl:creditscoring_age}).

\begin{table}[h!]
\caption{Credit scoring: performance of predictive models. Protected attribute: Sex}
\label{tbl:creditscoring_sex}
\begin{adjustbox}{width=\textwidth}
\begin{tabular}{l|l|c|c|c|c|c|c|c|c|c}
\hline
\textbf{Classifier} & \textbf{Method} & \textbf{Acc} & \textbf{BA} & \textbf{SP} & \textbf{EO} & \textbf{EOd} & \textbf{PP} & \textbf{PE} & \textbf{TE} & \textbf{ABROCA} \\
\hline
\multirow{5}{*}{MLP}
& Original &\textbf{\textcolor{black}{0.9980}} &\textbf{\textcolor{black}{0.9891}} &\textit{0.0305}&\textbf{\textcolor{black}{0.0006}} &\textbf{\textcolor{black}{0.0218}} &\textbf{\textcolor{black}{0.0007}} &\textbf{\textcolor{black}{0.0212}}&\textbf{\textcolor{black}{0.0}} &\textbf{\textcolor{black}{0.0095}} \\
& CTGAN &0.9011 & \textit{0.8597} & 0.1358&0.0650&0.1918&0.0074&0.1269&1.3611&\textit{0.0332}\\
& TabFairGan &\textit{0.9179}&0.6419&0.0477&0.0237&0.1692&0.0368&0.1455&0.1660&0.0784\\
& FairFinGAN-SP &0.9064&0.5430&-0.0394&0.0280&0.1429&0.0398&0.1150&-0.2089&0.0351\\
& FairFinGAN-EOd &0.9107&0.6017&\textbf{\textcolor{black}{0.0094}}&\textit{0.0084}&\textit{0.0670}&\textit{0.0044}&\textit{0.0586}&\textit{0.1190}&0.0624\\
\hline
\multirow{5}{*}{kNN}
& Original &\textbf{\textcolor{black}{0.9809}} &\textbf{\textcolor{black}{0.9108}} &0.0324&\textit{0.0061}&\textbf{\textcolor{black}{0.0626}}&\textbf{\textcolor{black}{0.0044}} &\textit{0.0566}&0.3963&\textbf{\textcolor{black}{0.0200}} \\
& CTGAN &0.8103&0.5890&0.0667&0.0543&0.0963&0.0753&\textbf{\textcolor{black}{0.0420}}&0.1817&0.0453\\
& TabFairGan &\textit{0.9212}&\textit{0.6140}&0.0384&0.0158&0.1866&0.0385&0.1708&0.1076&0.0731\\
& FairFinGAN-SP &0.9124&0.5436&\textit{-0.0252}&0.0132&0.1158&0.0395&0.1027&\textit{-0.0601}&0.0726\\
& FairFinGAN-EOd &0.9162&0.5826&\textbf{\textcolor{black}{0.0019}}&\textbf{\textcolor{black}{0.0057}}&\textit{0.0631}&\textit{0.0067}&0.0574&\textbf{\textcolor{black}{0.0465}}&\textit{0.0433}\\
\hline
\multirow{5}{*}{DT}
& Original &\textbf{\textcolor{black}{0.9983}} &\textbf{\textcolor{black}{0.9899}} &\textit{0.0303}&\textbf{\textcolor{black}{0.0002}} &\textbf{\textcolor{black}{0.0151}} &\textbf{\textcolor{black}{0.0}} &\textbf{\textcolor{black}{0.0149}} &\textbf{\textcolor{black}{0.0}} &\textbf{\textcolor{black}{0.0076}} \\
& CTGAN &0.8702&\textit{0.7965}&0.0804&0.0268&0.0730&0.0320&0.0462&-0.1304&0.0230\\
& TabFairGan &\textit{0.8778}&0.6398&0.0639&0.0416&0.1814&0.0370&0.1398&-0.0896&0.0619\\
& FairFinGAN-SP &0.8512&0.5617&-0.0530&0.0448&0.1194&0.0421&0.0746&0.0912&0.0248\\
& FairFinGAN-EOd &0.8660&0.6148&\textbf{\textcolor{black}{-0.0008}} &\textit{0.0088}&\textit{0.0534}&\textit{0.0069}&\textit{0.0446}&\textit{-0.0459}&\textit{0.0191}\\
\hline
\multirow{5}{*}{LR}
& Original &\textbf{\textcolor{black}{0.9959}} &\textbf{\textcolor{black}{0.9893}} &0.0313&\textbf{\textcolor{black}{0.0014}}&\textbf{\textcolor{black}{0.0183}}&\textbf{\textcolor{black}{0.0002}} &\textbf{\textcolor{black}{0.0169}}&\textbf{\textcolor{black}{0.0}} &\textbf{\textcolor{black}{0.0080}} \\
& CTGAN &0.8973&\textit{0.8083}&0.0935&0.0390&0.0919&0.0320&\textit{0.0529}&0.2784&\textit{0.0356}\\
& TabFairGan &\textit{0.9309}&0.6403&0.0420&0.0146&0.2118&0.0341&0.1971&0.1680&0.0451\\
& FairFinGAN-SP &0.9189&0.5435&\textit{-0.0244}&0.0111&0.1252&0.0385&0.1141&-0.0865&0.0364\\
& FairFinGAN-EOd &0.9247&0.6186&\textbf{\textcolor{black}{0.0045}}&\textit{0.0080}&\textit{0.0745}&\textit{0.0053}&0.0664&\textit{0.0724}&0.0546\\
\hline
\end{tabular}
\end{adjustbox}
\end{table}

\begin{table}[h!]
\caption{Credit scoring: performance of predictive models. Protected attribute: Age}
\label{tbl:creditscoring_age}
\begin{adjustbox}{width=\textwidth}
\begin{tabular}{l|l|c|c|c|c|c|c|c|c|c}
\hline
\textbf{Classifier} & \textbf{Method} & \textbf{Acc} & \textbf{BA} & \textbf{SP} & \textbf{EO} & \textbf{EOd} & \textbf{PP} & \textbf{PE} & \textbf{TE} & \textbf{ABROCA} \\
\hline
\multirow{5}{*}{MLP}
& Original &\textbf{\textcolor{black}{0.9979}} &\textbf{\textcolor{black}{0.9898}} &0.0814&\textbf{\textcolor{black}{0.0035}} &\textbf{\textcolor{black}{0.0291}} &\textbf{\textcolor{black}{0.0080}} &\textbf{\textcolor{black}{0.0256}}&\textbf{\textcolor{black}{0.0}} &\textbf{\textcolor{black}{0.0087}} \\
& CTGAN &0.8917&\textit{0.7734}&\textit{0.0581}&0.0199&0.1016&\textit{0.0257}&0.0818&0.2029&\textit{0.0328}\\
& TabFairGan &0.9285&0.6398&0.1069&0.0580&0.2591&0.0547&0.2011&0.3632&0.0750\\
& FairFinGAN-SP &\textit{0.9349}&0.5112&\textbf{\textcolor{black}{0.0124}} &\textit{0.0126}&\textit{0.0391}&0.0336&\textit{0.0265}&\textit{0.0945}&0.0751\\
& FairFinGAN-EOd &0.9022&0.6272&0.0888&0.0652&0.1545&0.0625&0.0893&0.3528&0.0747\\
\hline
\multirow{5}{*}{kNN}
& Original &\textbf{\textcolor{black}{0.9809}} &\textbf{\textcolor{black}{0.9108}} &0.0817&\textbf{\textcolor{black}{0.0068}}&0.0861&\textbf{\textcolor{black}{0.0155}} &0.0793&\textit{0.0548}&\textbf{\textcolor{black}{0.0267}} \\
& CTGAN &0.8103&0.5890&\textit{0.0240}&0.0209&\textit{0.0823}&0.0539&\textit{0.0615}&-0.0785&0.0507\\
& TabFairGan &0.9310&\textit{0.6204}&0.0682&0.0245&0.1822&0.0623&0.1577&0.0550&0.0564\\
& FairFinGAN-SP &\textit{0.9342}&0.5097&\textbf{\textcolor{black}{0.0069}} &\textit{0.0082}&\textbf{\textcolor{black}{0.0237}}&\textit{0.0337}&\textbf{\textcolor{black}{0.0154}}&\textbf{\textcolor{black}{0.0139}}&\textit{0.0292}\\
& FairFinGAN-EOd &0.9029&0.5874&0.0471&0.0412&0.1106&0.0765&0.0694&0.1130&0.0479\\
\hline
\multirow{5}{*}{DT}
& Original &\textbf{\textcolor{black}{0.9982}} &\textbf{\textcolor{black}{0.9892}} &\textit{0.0792}&\textbf{\textcolor{black}{0.0001}} &\textbf{\textcolor{black}{0.0276}} &\textbf{\textcolor{black}{0.0081}} &\textbf{\textcolor{black}{0.0275}} &\textbf{\textcolor{black}{0.0}} &\textbf{\textcolor{black}{0.0138}} \\
& CTGAN &0.8717&\textit{0.7988}&0.0796&\textit{0.0387}&\textit{0.1062}&\textit{0.0202} &\textit{0.0675}&0.2560&0.0368\\
& TabFairGan &\textit{0.8866}&0.6229&0.0951&0.0583&0.2334&0.0646&0.1751&-0.1897&0.0733\\
& FairFinGAN-SP &0.8834&0.5425&\textbf{\textcolor{black}{0.0599}}&0.0565&0.1557&0.0296&0.0992&0.3557&0.0379\\
& FairFinGAN-EOd &0.8660&0.6370&0.0943&0.0725&0.1422&0.0626&0.0697&\textit{0.0523}&\textit{0.0337}\\
\hline
\multirow{5}{*}{LR}
& Original &\textbf{\textcolor{black}{0.9959}} &\textbf{\textcolor{black}{0.9893}} &0.0809&\textit{0.0049}&\textit{0.0321}&\textbf{\textcolor{black}{0.0081}} &\textit{0.0272}&\textbf{\textcolor{black}{0.0}} &\textbf{\textcolor{black}{0.0135}} \\
& CTGAN &0.8973&\textit{0.8083}&0.0708&0.0175&0.0992&\textit{0.0185}&0.0817&0.1921&\textit{0.0355}\\
& TabFairGan &0.9399&0.6338&0.0806&0.0242&0.2503&0.0502&0.2261&0.1731&0.0451\\
& FairFinGAN-SP &\textit{0.9400}&0.5040&\textbf{\textcolor{black}{0.0038}} &\textbf{\textcolor{black}{0.0033}} &\textbf{\textcolor{black}{0.0176}} &0.0326&\textbf{\textcolor{black}{0.0143}} &\textit{0.0221}&0.0448\\
& FairFinGAN-EOd &0.9172&0.6040&\textit{0.0629}&0.0345&0.1614&0.0594&0.1269&0.2132&0.0657\\
\hline
\end{tabular}
\end{adjustbox}
\end{table}

Table \ref{tbl:dutch_sex} presents the results on the Dutch census dataset. Across all classifiers, FairFinGAN-EOd achieves the highest accuracy while maintaining good fairness performance (EO, EOd, PP). In contrast, FairFinGAN-SP reduces accuracy, but yields the smallest deviations on fairness measures, especially SP across all classifiers. CTGAN and TabFairGAN often introduce notable drops in accuracy.
In the German credit dataset (Table \ref{tbl:german_sex}), FairFinGAN-SP produces good results in terms of fairness (SP, EO, EOd, and ABROCA), especially for LR and DT classifiers, although sometimes with reduced accuracy. FairFinGAN-EOd improves fairness in the MLP and DT classifiers. Interestingly, CTGAN often yields the highest accuracy and fairness (EO, EOd) across all classifiers.

\begin{table}[h!]
\caption{Dutch census: performance of predictive models. Protected attribute: Sex}
\label{tbl:dutch_sex}
\begin{adjustbox}{width=\textwidth}
\begin{tabular}{l|l|c|c|c|c|c|c|c|c|c}
\hline
\textbf{Classifier} & \textbf{Method} & \textbf{Acc} & \textbf{BA} & \textbf{SP} & \textbf{EO} & \textbf{EOd} & \textbf{PP} & \textbf{PE} & \textbf{TE} & \textbf{ABROCA} \\
\hline
\multirow{5}{*}{MLP}
& Original &\textit{0.8021}&\textit{0.8036}&0.3565&0.0945&0.3732&0.0382&0.2788&0.2509&\textit{0.0307}\\
& CTGAN &0.8017&0.7919&\textit{0.2750}&0.1467&0.2874&0.0556&\textbf{\textcolor{black}{0.1407}}&0.7251&\textbf{\textcolor{black}{0.0186}} \\
& TabFairGan &0.7682&0.7657&0.3599&0.2106&0.4634&\textit{0.0312}&0.2528&1.2062&0.0407\\
& FairFinGAN-SP &0.6145&0.6231&\textbf{\textcolor{black}{0.1805}}&\textit{0.0608}&\textbf{\textcolor{black}{0.2330}}&0.1948&\textit{0.1722}&\textbf{\textcolor{black}{0.1398}} &0.1004\\
& FairFinGAN-EOd &\textbf{0.8513}&\textbf{\textcolor{black}{0.8522}} &0.3340&\textbf{\textcolor{black}{0.0445}} &\textit{0.2773}&\textbf{\textcolor{black}{0.0234}} &0.2328&\textit{0.2323}&0.0476\\
\hline
\multirow{5}{*}{kNN}
& Original &\textit{0.8236}&\textit{0.8226}&0.3250&0.0759&0.2834&\textbf{\textcolor{black}{0.0131}} &0.2076&0.4582&0.0475\\
& CTGAN &0.8145&0.8074&\textit{0.2361}&0.0695&0.1797&0.0715&\textit{0.1103} &\textit{-0.0291}&\textit{0.0217}\\
& TabFairGan &0.7901&0.7896&0.3048&0.1154&0.3142&0.0592&0.1987&0.3170&0.0401\\
& FairFinGAN-SP &0.4934&0.4934&\textbf{\textcolor{black}{-0.0016}} &\textbf{\textcolor{black}{0.0081}} &\textbf{\textcolor{black}{0.0176}} &0.2856&\textbf{\textcolor{black}{0.0094}} &-0.8618&\textbf{\textcolor{black}{0.0041}} \\
& FairFinGAN-EOd &\textbf{\textcolor{black}{0.8659}} &\textbf{\textcolor{black}{0.8657}} &0.2882&\textit{0.0342}&\textit{0.1648}&\textit{0.0333}&0.1306&\textbf{\textcolor{black}{0.0072}} &0.0379\\
\hline
\multirow{5}{*}{DT}
& Original &\textit{0.8152}&\textit{0.8139}&0.3359&0.0959&0.3189&\textbf{\textcolor{black}{0.0106}} &0.2229&0.6652&0.0593\\
& CTGAN &0.8008&0.7922&\textit{0.2446}&\textit{0.0832}&\textit{0.2124}&0.0624&\textbf{\textcolor{black}{0.1291}}&\textbf{\textcolor{black}{0.0966}} &\textit{0.0364}\\
& TabFairGan &0.7906&0.7890&0.2887&0.1093&0.2815&0.0618&0.1722&0.2654&\textbf{\textcolor{black}{0.0390}}\\
& FairFinGAN-SP &0.6535&0.6537&\textbf{\textcolor{black}{0.1381}}&0.0638&0.2877&0.1720&0.2239&-0.7600&0.1019\\
& FairFinGAN-EOd &\textbf{\textcolor{black}{0.8609}}&\textbf{\textcolor{black}{0.8604}} &0.2962&\textbf{\textcolor{black}{0.0461}} &\textbf{\textcolor{black}{0.1889}}&\textit{0.0236}&\textit{0.1429}&\textit{0.1775}&0.0465\\
\hline
\multirow{5}{*}{LR}
& Original &\textit{0.8160}&\textit{0.8148}&0.3271&0.0908&0.2986&\textbf{\textcolor{black}{0.0194}} &0.2078&0.4886&\textit{0.0302}\\
& CTGAN &0.8159&0.8065&\textit{0.2155}&\textit{0.0641}&\textit{0.1489}&0.0846&\textit{0.0848}& \textbf{\textcolor{black}{-0.2971}}&\textbf{\textcolor{black}{0.0205}} \\
& TabFairGan &0.7325&0.7306&0.2574&0.1097&0.2821&0.1183&0.1725&\textit{-0.1608}&0.0330\\
& FairFinGAN-SP &0.7574&0.7562&\textbf{\textcolor{black}{0.1204}}&0.1518&0.2297&0.1198&\textbf{\textcolor{black}{0.0779}}&-1.6286&0.1241\\
& FairFinGAN-EOd &\textbf{\textcolor{black}{0.8415}}&\textbf{\textcolor{black}{0.8400}} &0.2483&\textbf{\textcolor{black}{0.0288}} &\textbf{\textcolor{black}{0.1381}} &\textit{0.0470}&0.1094&-0.5412&0.0536\\
\hline
\end{tabular}
\end{adjustbox}
\end{table}

\begin{table}[h!]
\caption{German credit: performance of predictive models. Protected attribute: Sex}
\label{tbl:german_sex}
\begin{adjustbox}{width=\textwidth}
\begin{tabular}{l|l|c|c|c|c|c|c|c|c|c}
\hline
\textbf{Classifier} & \textbf{Method} & \textbf{Acc} & \textbf{BA} & \textbf{SP} & \textbf{EO} & \textbf{EOd} & \textbf{PP} & \textbf{PE} & \textbf{TE} & \textbf{ABROCA} \\
\hline
\multirow{5}{*}{MLP}
& Original &0.4830&0.4974&\textit{-0.0200}&0.0215&0.0456&0.1651&0.0241&-1.9444&0.0867\\
& CTGAN &\textbf{\textcolor{black}{0.7784}} &0.5017&0.0194&\textbf{\textcolor{black}{0.0199}}&\textit{0.0370}&\textit{0.0240}&\textit{0.0172}&\textbf{\textcolor{black}{0.0858}}&\textbf{\textcolor{black}{0.0216}} \\
& TabFairGan &\textit{0.7300}&\textbf{\textcolor{black}{0.5143}}&-0.0270&0.0281&0.0559&0.0704&0.0278&\textit{-0.1697}&0.0533\\
& FairFinGAN-SP &0.5774&0.5072&0.0413&0.0449&0.0758&\textbf{\textcolor{black}{0.0218}} &0.0309&0.5652&\textit{0.0285}\\
& FairFinGAN-EOd &0.5570&\textit{0.5107}&\textbf{\textcolor{black}{-0.0167}}&\textit{0.0210}&\textbf{\textcolor{black}{0.0307}}&0.0541&\textbf{\textcolor{black}{0.0098}}&-12.2333&0.0878\\
\hline
\multirow{5}{*}{kNN}
& Original &0.6660&\textit{0.5376}&\textbf{\textcolor{black}{0.0004}} &0.0161&0.0312&0.0766&0.0151&-0.1756&0.0786\\
& CTGAN &\textbf{\textcolor{black}{0.7851}} &0.5022&0.0102&\textbf{\textcolor{black}{0.0122}} &\textbf{\textcolor{black}{0.0201}} &\textit{0.0252}&\textbf{\textcolor{black}{0.0079}} &0.0238&\textbf{\textcolor{black}{0.0184}} \\
& TabFairGan &\textit{0.7150}&\textbf{\textcolor{black}{0.5412}}&-0.0116&0.0245&0.1302&0.0553&0.1058&0.1195&0.0930\\
& FairFinGAN-SP &0.6761&0.4976&\textit{-0.0018}&\textit{0.0130}&\textit{0.0279}&\textbf{\textcolor{black}{0.0173}} &\textit{0.0149}&\textit{-0.0197}&\textit{0.0209}\\
& FairFinGAN-EOd &0.6020&0.5018&0.0077&0.0191&0.0963&0.0520&0.0772&\textbf{\textcolor{black}{-0.0126}} &0.0828\\
\hline
\multirow{5}{*}{DT}
& Original &\textbf{\textcolor{black}{0.6870}}&\textbf{\textcolor{black}{0.6307}} &0.0224&0.0626&0.1565&0.0576&0.0939&-0.3367&0.0776\\
& CTGAN &\textit{0.6824}&\textit{0.5148}&0.0270&\textit{0.0269}&\textit{0.0572}&\textit{0.0235}& \textit{0.0303}&\textbf{\textcolor{black}{0.0007}} &\textit{0.0110}\\
& TabFairGan &0.6010&0.5011&-0.0524&0.0972&0.1782&0.0758&0.0810&0.1844&0.0666\\
& FairFinGAN-SP &0.6025&0.5041&\textit{0.0112}&\textbf{\textcolor{black}{0.0160}} &\textbf{\textcolor{black}{0.0323}} &\textbf{\textcolor{black}{0.0162}} &\textbf{\textcolor{black}{0.0164}} &\textit{0.0052}&\textbf{\textcolor{black}{0.0040}} \\
& FairFinGAN-EOd &0.5720&0.5147&\textbf{\textcolor{black}{0.0002}} &0.0583&0.1825&0.0815&0.1242&-0.0229&0.0721\\
\hline
\multirow{5}{*}{LR}
& Original &0.7210&\textbf{\textcolor{black}{0.6017}}&0.0453&0.0345&0.1831&0.0585&0.1486&-0.0999&0.0857\\
& CTGAN &\textbf{\textcolor{black}{0.8125}} &0.5&\textbf{\textcolor{black}{0.0}}&\textbf{\textcolor{black}{0.0}}&\textbf{\textcolor{black}{0.0}}&\textit{0.0234}&\textbf{\textcolor{black}{0.0}}&\textbf{\textcolor{black}{0.0}}&\textbf{\textcolor{black}{0.0298}}\\
& TabFairGan &\textit{0.7420}&0.5013&\textit{0.0013}&\textit{0.0022}&\textit{0.0147}&0.0725&\textit{0.0125}&\textit{-0.0050}&0.0907\\
& FairFinGAN-SP &0.7346&0.5&\textbf{\textcolor{black}{0.0}} &\textbf{\textcolor{black}{0.0}} &\textbf{\textcolor{black}{0.0}} &\textbf{\textcolor{black}{0.0165}} &\textbf{\textcolor{black}{0.0}} &\textbf{\textcolor{black}{0.0}} &\textit{0.0350}\\
& FairFinGAN-EOd & 0.6700 &\textit{0.5106} & 0.0083 & 0.0319 & 0.0900 & 0.0637 & 0.0581 & 0.0529 & 0.0965\\
\hline
\end{tabular}
\end{adjustbox}
\end{table}

In summary, the variation in fairness gains between predictive models suggests that the effectiveness of the proposed synthetic data approach depends on the underlying learning mechanism, underscoring the need for model-specific evaluation. In financial applications such as lending or credit scoring, fairness-aware data generation can help reduce historical bias, promote more equitable risk assessment, and better align automated decisions with regulatory requirements while preserving predictive performance.


\section{Conclusions and Outlook}
\label{sec:conclusion}

In this paper, we proposed FairFinGAN, a fairness-aware WGAN-based framework for generating synthetic financial datasets. By integrating a fairness score with a classifier into the GAN training process, our approach can address bias at the dataset level rather than at the classification level. Experimental results on five real-world financial datasets demonstrated that FairFinGAN effectively mitigates bias while maintaining data utility, outperforming existing GAN-based data generation methods. In the future, we plan to extend FairFinGAN to handle multiple protected attributes simultaneously and explore its application to other domains such as healthcare and education. Additionally, investigating more advanced fairness metrics and incorporating differential privacy could further enhance the reliability and applicability of the generated data.

%
%
%
\bibliographystyle{splncs04}
\bibliography{bibliography}
\end{document}